\documentclass[11pt,a4paper]{article}
\usepackage[hyperref]{acl2017}
\usepackage{times}
\usepackage{latexsym}
\usepackage{amsmath}
\usepackage{amssymb}
\usepackage{hyphenat}

\usepackage{graphicx}
\usepackage{subfig}
\usepackage{tikz}
\usepackage{pgfplots} \pgfplotsset{compat=1.14}
\usepackage{multirow}

\usepackage{booktabs}
\usepackage{cleveref}
\usepackage{url}
\usepackage[multiple]{footmisc}

\usepackage{todonotes}

\newcommand\feat[1]{\texttt{\small #1}}

\makeatletter
\def\blfootnote{\xdef\@thefnmark{}\@footnotetext}
\makeatother

\aclfinalcopy % Uncomment this line for the final submission
\title{Are distributional representations ready for the real world?\\Evaluating word vectors for grounded perceptual meaning}

\author{
Li Lucy$^{1}$\\
\texttt{lucy3@stanford.edu}\\
\And
Jon Gauthier$^{1,2}$\\
\texttt{jon@gauthiers.net}\AND\\[-5ex]
{$^{1}$Stanford Symbolic Systems\quad
$^{2}$Stanford NLP Group\quad}}

\date{}

\begin{document}
\maketitle
\begin{abstract}
Distributional word representation methods exploit word co-occurrences to build compact vector encodings of words. While these representations enjoy widespread use in modern natural language processing, it is unclear whether they accurately encode all necessary facets of conceptual meaning. In this paper, we evaluate how well these representations can predict perceptual and conceptual features of concrete concepts, drawing on two semantic norm datasets sourced from human participants. We find that several standard word representations fail to encode many salient perceptual features of concepts, and show that these deficits correlate with word-word similarity prediction errors. Our analyses provide motivation for grounded and embodied language learning approaches, which may help to remedy these deficits.\blfootnote{All project code available at \href{https://github.com/lucy3/Graphs-Embeddings/tree/robonlp}{\path{github.com/lucy3/grounding-embeddings}}.}
\end{abstract}

\section{Introduction} % INTRODUCTION

Distributional approaches to meaning representation have enabled a substantial amount of progress in natural language processing over the past years. They center around a classic insight from at least as early as \citet{harris1954distributional,firth1957synopsis}:
\begin{quote}You shall know a word by the company it keeps. \citep[p. 11]{firth1957synopsis}\end{quote}

Popular distributional analysis methods which exploit this intuition such as word2vec \citep{mikolov2013distributed} and GloVe \citep{pennington2014glove} have been critical to the success of many recent large-scale natural language processing applications \citep[e.g.][]{turney2010frequency,turian2010word,collobert2008unified,socher2013recursive,goldberg2016primer}. These methods operationalize distributional meaning via tasks where words are optimized to predict words which co-occur with them in text corpora. These methods yield compact word representations --- vectors in some high-dimensional space --- which are optimized to solve these prediction tasks. These vector representations form the foundation of practically all modern deep learning models applied within natural language processing.

Despite the success of distributional representations in standard natural language processing tasks, a small but growing consensus within the artificial intelligence community suggests that these methods cannot be sufficient to induce adequate representations of words and concepts \citep{kiela2016virtual,gauthier2016paradigm,Lazaridou2015CombiningLA}. These sorts of claims, which often draw on experimental evidence from cognitive science \citep[see e.g.][]{barsalou2008grounded}, are used to back up arguments for multimodal learning (at the weakest) or complete embodiment (at the strongest). \citet{kiela2016virtual} claim the following:
\begin{quote}\ldots the best way for acquiring human-level semantics is to have machines learn through (physical) experience: if we want to teach a system the true meaning of ``bumping into a wall,'' we simply have to bump it into walls repeatedly.\end{quote}

Discussions like the one above have an intuitive pull: certainly ``bump'' is best understood through a sense of touch, just as ``loud'' is best understood through a sense of sound. It seems inefficient --- or perhaps just wrong --- to learn these sorts of concepts from distributional evidence.

Despite the intuitive pull, there is not much evidence from a computational perspective that grounded or multimodal learning actually earns us anything in terms of general meaning representation. Will our robots and chat-bots be worse off for not having physically bumped into walls before they hold discussions on wall-collisions? Will our representation of the concept \emph{loud} somehow be faulty unless we explicitly associate it with certain decibel levels experienced in the real world? Before we proceed to embed our learning agents in multimodal games and robot-shells, it is important that we have some concrete idea of how grounding actually affects meaning.

This paper presents a thorough analysis of the contents of distributional word representations with respect to this question. Our results suggest that several common distributional word representations may indeed be deficient in the sort of grounded meaning necessary for language-enabled agents deployed in the real world.

\section{Related work} \label{rel_work}% RELATED WORK

This paper uses semantic norm datasets to evaluate the content of distributional word representations. Semantic norm datasets consist of concepts and norms concerning their perceptual and conceptual features, as provided by human participants. They are a popular resource within psychology and cognitive science as models of human concept representation, and have been used to explain psycholinguistic phenomena from semantic priming and interference \citep{vigliocco2004representing} to the structure of early word learning in child language acquisition \citep{hills2009categorical}. \citet{andrews2009integrating} show how ``experiential'' semantic norm information can be used to model human judgments of concept similarity. They show that this semantic norm data provides information distinct from the information found in basic word representations. Our work extends the findings of \citeauthor{andrews2009integrating} to a larger semantic norm dataset and evaluates particular implications within natural language processing.

A small NLP literature has compared distributional representations with semantic norm datasets and other external resources. \citet{rubinstein_how_2015} confirm that word representations are especially effective at predicting taxonomic features versus attributive features. \citet{collell_is_2016} find that word representations fail to predict many visual features of concepts, and show how representations from computer vision models can help improve these predictions. Several studies have used distributional representations to reconstruct aspects of these semantic norm datasets \citep{Herbelot2015BuildingAS,fagarasan_distributional_2015,erk-2016-alligator}.

The majority of the NLP work in this space has focused on the downstream task of augmenting word representations with novel grounded information, often evaluating on standard semantic similarity datasets \citep{agirre_study_2009,bruni_distributional_2012,faruqui_retrofitting_2015,Bulat2016VisionAF}. \citet{young2014image} develop an alternative operationalization of denotational meaning using image captioning datasets, and demonstrate gains over distributional representations on textual similarity and entailment datasets.

This applied work has demonstrated that \emph{something} worthwhile is indeed gained by augmenting distributional representations with some orthogonal grounded or multimodal information. We believe it is critical to analyze the original successes and failures of distributional representations in order to motivate this move to grounded meaning representation.

\section{Meaning representations}

\subsection{Distributional meaning}  % DIST MEANING
\label{dist_meaning}

\begin{table} % TABLE 1
\centering
\resizebox{\linewidth}{!}{
\begin{tabular}{l|cc}
\toprule
 & \# word tokens & \# word types \\
\midrule
GloVe (Common Crawl) & 840B & 2.2M\\
GloVe (Wiki+Gigaword) & 6B & 400K\\
word2vec & 100B & 3M \\
\bottomrule
\end{tabular}
}
\caption{Statistics of the corpora used to produce the distributional representations used in this paper.}
\label{tbl:distributional-norm-datasets}
\end{table}

This paper examines representations produced by two popular unsupervised distributional methods. \Cref{tbl:distributional-norm-datasets} shows the statistics of the corpora used to generate these vectors.

\subparagraph{GloVe:} GloVe \citep{pennington2014glove} estimates word representations $w_i$ by using them to reconstruct a word-word co-occurrence matrix $X$ collected from a large text corpus:
\begin{equation}
L = \sum_{i,j=1}^V f(X_{ij}) \left(w_i^T w_j + b_i + b_j - \log X_{ij}\right)^2
\end{equation}
here $f(X_{ij})$ is a weighting function on word pairs and $b_i, b_j$ are learned per-word bias terms.

We use two pre-trained GloVe vector datasets: one trained on a concatenation of Wikipedia 2014 and Gigaword 5 (GloVe-WG), and another trained on a Common Crawl dump (GloVe-CC).\footnote{\href{https://nlp.stanford.edu/projects/glove/}{\path{nlp.stanford.edu/projects/glove}}}

\subparagraph{word2vec:} word2vec \citep{mikolov2013distributed} estimates word representations by optimizing a skip-gram objective to predict all words $w_j$ within a context window $c$ of a word $w_i$ given their word representations:
\begin{equation}
J = \frac{1}{T} \sum_{i=1}^T \sum_{i-c \le j \le i+c} \log p(w_j \mid w_i)
\end{equation}
where $T$ is the total number of words in a corpus. We use a publicly available word2vec dataset trained on the Google News corpus.\footnote{\href{https://code.google.com/archive/p/word2vec/}{\path{code.google.com/archive/p/word2vec}}}

\subsection{Semantic norms} \label{norms} % SEMANTIC NORMS

\begin{table} % TABLE 2
\centering
\resizebox{\linewidth}{!}{
\begin{tabular}{l|cc|cc}
\toprule
Dataset & \# concepts & \# features & C/F & F/C \\
\midrule
McRae & 541 & 2526 & 2.87 & 13.41 \\
CSLB & 638 & 2725 & 3.78 & 16.13 \\
\bottomrule
\end{tabular}
}
\caption{Semantic norm datasets used in this paper. The final two columns show the mean concepts per feature / features per concept.}
\label{tbl:semantic-norm-datasets}
\end{table}

Semantic feature norm datasets consist of reports from human participants about the semantic features of various natural kinds. A proportion of the features contained in these datasets are properties of concepts which may be obvious to humans but are perhaps difficult to find written in text corpora. For this reason, we selected two semantic norm datasets to serve as gold-standard comparisons of concept meaning. \Cref{tbl:semantic-norm-datasets} displays basic statistics about the semantic norm datasets we use in this paper.

\paragraph{McRae} Our initial experiments use the semantic norm dataset from \citet{mcrae2005semantic}, which consists of 541 concrete noun concepts with associated feature norms, collected from 725 participants. For a given concept, the McRae dataset includes all feature norms which were reported independently by at least five participants (2,526 in total). After removing concepts indicated to have ambiguous meanings to mitigate polysemy effects (such as \texttt{tank\_(army)} and \texttt{tank\_(container)}) and one concept without a GloVe representation (\texttt{dunebuggy}), we had a resulting set of 515 concepts for analysis. The dataset groups features into several perceptual and non-perceptual categories: taxonomic, encyclopedic, function, visual-motion, visual-form\_and\_surface, visual-colour, sound, tactile, and taste \citep{mcrae2005semantic}. We use the McRae dataset and feature categories to perform basic pilot analyses and form hypotheses about the nature of the distributional representations tested.

\paragraph{CSLB} We reproduce and extend our results on a second semantic norm dataset collected by the Cambridge Centre for Speech, Language and the Brain~\citep[CSLB; ][]{devereux2014centre}. CSLB contains 638 concepts provided by 123 participants. Their data collection closely followed \citet{mcrae2005semantic}, though features were included if at least 2 participants named that feature. We removed concepts with two-word names, ambiguous meanings, or missing vector representations to yield a vocabulary of 597 concepts from this dataset. CSLB also includes a feature categorization schema, though the categories are broader than those in McRae: visual perceptual, other perceptual, functional, taxonomic, and encyclopedic.

\paragraph{}The mapping between the two categorization schemes is far from perfect. While some perceptual features in McRae are categorized as perceptual features in CSLB, other features (e.g. those related to swimming, flying, eating) are reclassified as ``functional'' in CSLB. The two datasets disagree on abstract conceptual properties as well. For example, CSLB classifies \texttt{is\_for\_football} as a functional property, while McRae classifies the comparable feature \texttt{associated\_with\_football\_games} as encyclopedic.

The encyclopedic category is somewhat difficult to distinguish in both datasets. It is composed mainly of abstract factual features, but also contains attributive features such as \texttt{is\_cold\_blooded} and \texttt{does\_use\_electricity} as well as \texttt{is\_scary} and \texttt{is\_cool}.

Meanwhile, the functional category mixes features for behaviors associated with the concept (\texttt{does\_dive}) as well as functions that people perform on or with the concept (\texttt{is\_hit}). This classification system may need some readjustments to provide a clear understanding of what is perceptual and what is conceptual, and it may be that some features, such as \texttt{has\_a\_steering\_wheel}, are both.

Given the significant noise of this classification scheme, we focus our investigation on a single contrast between features in clearly perceptual categories (visual, tactile, sound, etc.) and non-perceptual categories (functional and taxonomic). Because the encyclopedic category contains an ambiguous mix of both sorts, we exclude it from our formal predictions later in the paper.

\section{The feature view} % FEATURE VIEW
\label{sec:feature-view}

We first investigate how well distributional word representations directly encode information about semantic norms.\footnote{The remainder of this paper describes a general analysis performed on both the McRae and CSLB datasets. We used McRae as a pilot dataset to form hypotheses, and checked these hypotheses on the CSLB dataset as a test set. All of the graphs and numbers reported in this paper correspond to results on CSLB.} For each feature in a semantic norm dataset, we construct a binary classification problem which predicts the presence or absence of the feature for each concept. Concretely, for each feature $f_i$ we have a label vector $y_i \in \{0, 1\}^{n_c}$, where $n_c$ is the total number of concepts in the dataset, and $y_{ij}$ is 1 when concept $j$ has feature $f_i$ and 0 otherwise. We build label vectors only for features with five or more associated concepts. After filtering, we have $n_f = 267$ label vectors in the McRae dataset and $n_f = 775$ in CSLB.

For each feature, we construct a binary logistic regression model $p^{i}$ which predicts the presence or absence of the feature for a concept given its word representation $x_j$:
\begin{equation}
p^{i}(y_{ij} \mid x_j) = \sigma(w_i^T x_j) \label{eqn:regression}
\end{equation}

This base model is extremely prone to overfitting, as most features have only several associated concepts --- that is, each classifier has only a few positive examples --- and the input word representations are of a high dimensionality. In order to prevent overfitting, we add an independent L2 regularization term to each regression model. For each feature $f_i$, we use leave-one-out cross-validation to select the regularization parameter $\lambda_i$ which maximizes the following modified logistic objective:
\begin{equation}
\begin{split}
L_i(\lambda_i) &=\, \frac{1}{|f_i|} \sum_{x_j\in f_i} \left( \log p^{i,\lambda_i}_{-j}(y_{ij} = 1 \mid x_j) \vphantom{\sum_{x_k \not \in f_i}}\right. \\
    & + \left.\frac{1}{n_c - |f_i|}\sum_{x_k \not \in f_i}\log p^{i,\lambda_i}_{-j}(y_{ik} = 0 \mid x_k) \right)%
\end{split}
\label{eqn:loocv}
\end{equation}

Here $p^{i,\lambda_i}_{-j}(\cdot)$ represents a regression model (\Cref{eqn:regression}) trained without example $(x_j, y_{ij})$ in the training set and with regularization parameter $\lambda_i$. The first term of the summand calculates the log-probability of the left-out concept having the desired feature, and the second term calculates the average log-probability that any other concept (outside of the feature group $f_i$) does not have the feature. The regularization terms $\lambda_i$ are selected independently for each feature to maximize the objective $L_i$.

\begin{figure*}[htb] % FIGURE 1
\centering
\includegraphics[clip,width=0.85\linewidth]{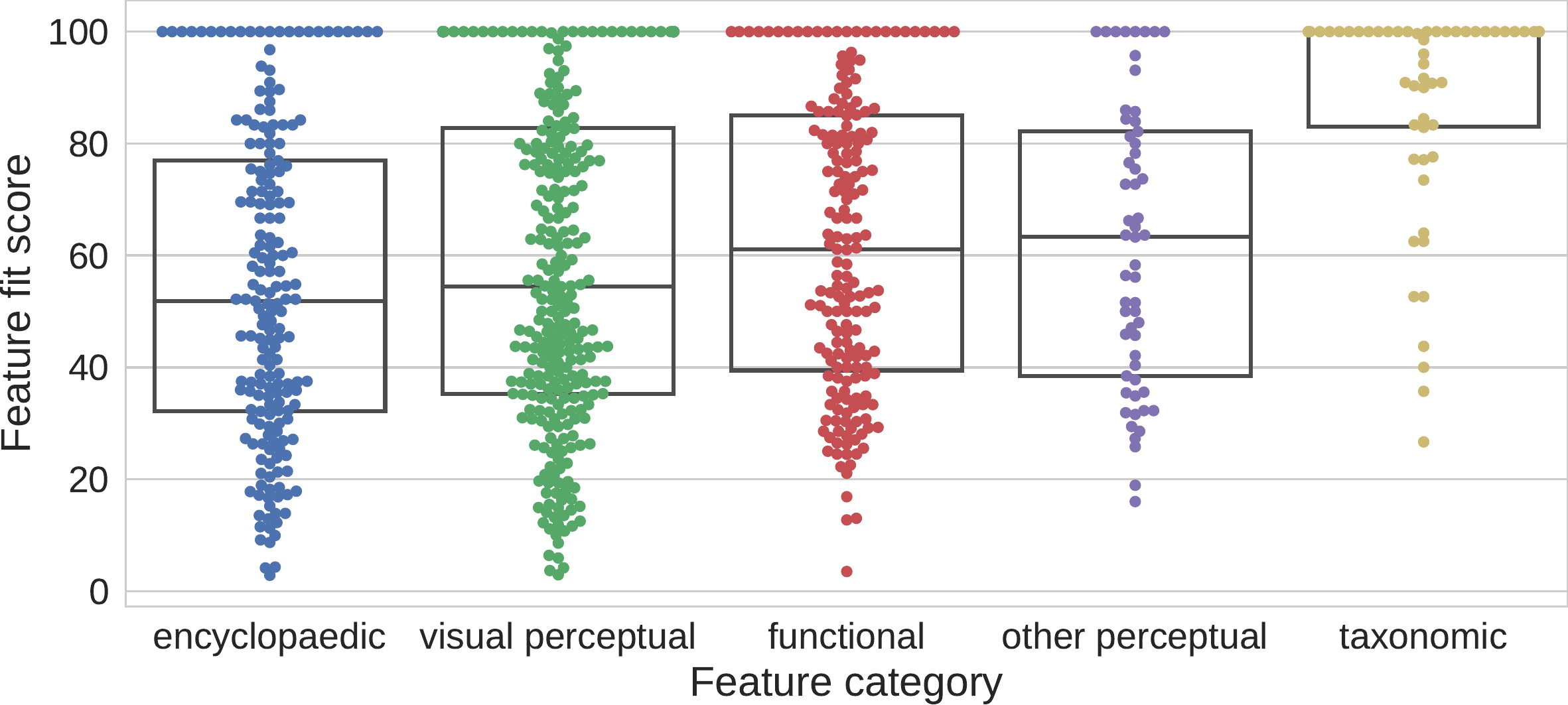}
\caption{The CSLB feature fit metrics of GloVe-CC, where each point is a feature with at least 5 associated concepts. Feature categories are on the horizontal axis.}
\label{fig:feature_fit}
\end{figure*}

After fitting the regularized logistic regression models, we calculate a set of ``feature fit'' metrics. For each feature $f_i$, we evaluate the binary F1 score of its classifier's predictions $p^{i}(y_i)$. \Cref{fig:feature_fit} shows each feature as a point in a swarm-plot (grouped by feature category).

\begin{table*}
\centering
\resizebox{\linewidth}{!}{
\begin{tabular}{p{2.8cm}|p{7cm}|p{7cm}}
\toprule
\textnormal{category} & \textnormal{feature fit $< 50\%$} & \textnormal{feature fit $> 50\%$} \\
\midrule
\textnormal{other perceptual} & \feat{is\_chewy}, \feat{is\_solid}, \feat{is\_high\_pitched} & \feat{is\_hard}, \feat{does\_smell\_good\_nice}, \feat{is\_juicy}\\
\textnormal{visual perceptual} & \feat{is\_triangular}, \feat{has\_a\_string}, \feat{is\_curved}, & \feat{has\_a\_clasp}, \feat{has\_a\_shell}, \feat{has\_whiskers} \\
\textnormal{encyclopedic} & \feat{is\_collectable}, \feat{is\_powerful}, \feat{made\_of\_tissue} & \feat{is\_formal}, \feat{does\_not\_fly}, \feat{is\_kept\_in\_a\_cage}\\
\textnormal{functional} & \feat{is\_roasted}, \feat{is\_for\_weddings}, \feat{is\_carried} & \feat{does\_shelter}, \feat{does\_chop}, \feat{is\_eaten\_edible}\\
\textnormal{taxonomic} & \feat{is\_a\_home}, \feat{is\_a\_vessel}, \feat{is\_an\_ingredient} & \feat{is\_seafood}, \feat{is\_a\_boat}, \feat{is\_a\_tool} \\
\bottomrule
\end{tabular}
}
\caption{Examples of features in each category with feature fit scores based on using GloVe-CC to predict norms from CSLB.}
\label{tbl:feature_examples}
\end{table*}

Pilot tests with the McRae dataset suggested that the categories associated with strictly perceptual features were not well encoded in the distributional representations relative to strictly non-perceptual categories (taxonomic and functional features).

We use the CSLB dataset as a test set for this prediction. We perform a bootstrap confidence interval test on the difference between the median feature fit scores for CSLB features in non-perceptual and perceptual categories. The 95\% confidence intervals on this bootstrap are positive for two of the three representations tested (GloVe-CC and word2vec).\footnote{GloVe-CC: (7.67\%, 24.0\%); word2vec: (7.13\%, 20.6\%); GloVe-WG: (-1.25\%, 15.7\%).} \Cref{fig:feature_fit} shows the feature fit scores on CSLB evaluated with GloVe-CC, and the word2vec evaluation effectively shows the same result: taxonomic and functional features score higher on average than strictly perceptual features. This comparison failed on GloVe-WG, however, where features classed as ``functional'' scored far lower on average than those in perceptual categories. Across all three sets of distributional representations, the median score of encyclopedic features was well below all other feature categories.

It is obvious from \Cref{fig:feature_fit} that each category contains a wide range of feature fit values. As discussed earlier in~\Cref{norms}, this categorization of features is far from perfect. Many of the lower-scoring features classed as ``encyclopedic'' are simple attributive features not deserving of the category label, such as \texttt{is\_fresh} and \texttt{is\_filling}. Many of the higher-scoring encyclopedic features seem genuinely encyclopedic, such as \texttt{is\_found\_on\_farms}; other high-scoring features are arguably ``functional,'' such as \texttt{does\_grow\_on\_trees}. Many of the higher scoring visual perceptual features state structural part-whole relations, such as \texttt{has\_legs} and \texttt{has\_an\_engine}.

\Cref{tbl:feature_examples} provides more examples of low- and high-scoring features in each category. Despite the rather noisy classification scheme used in this dataset, we still managed to find a regular trend in two of three evaluations, matching our expectations from prior pilot experiments. We believe that a revised classification scheme could help to demonstrate a clear difference between perceptual and non-perceptual features in all three datasets.

\subsection{Matching word representation sources}
\label{sec:feature-comp}

\begin{figure}[t]
\centering
\includegraphics[width=\linewidth]{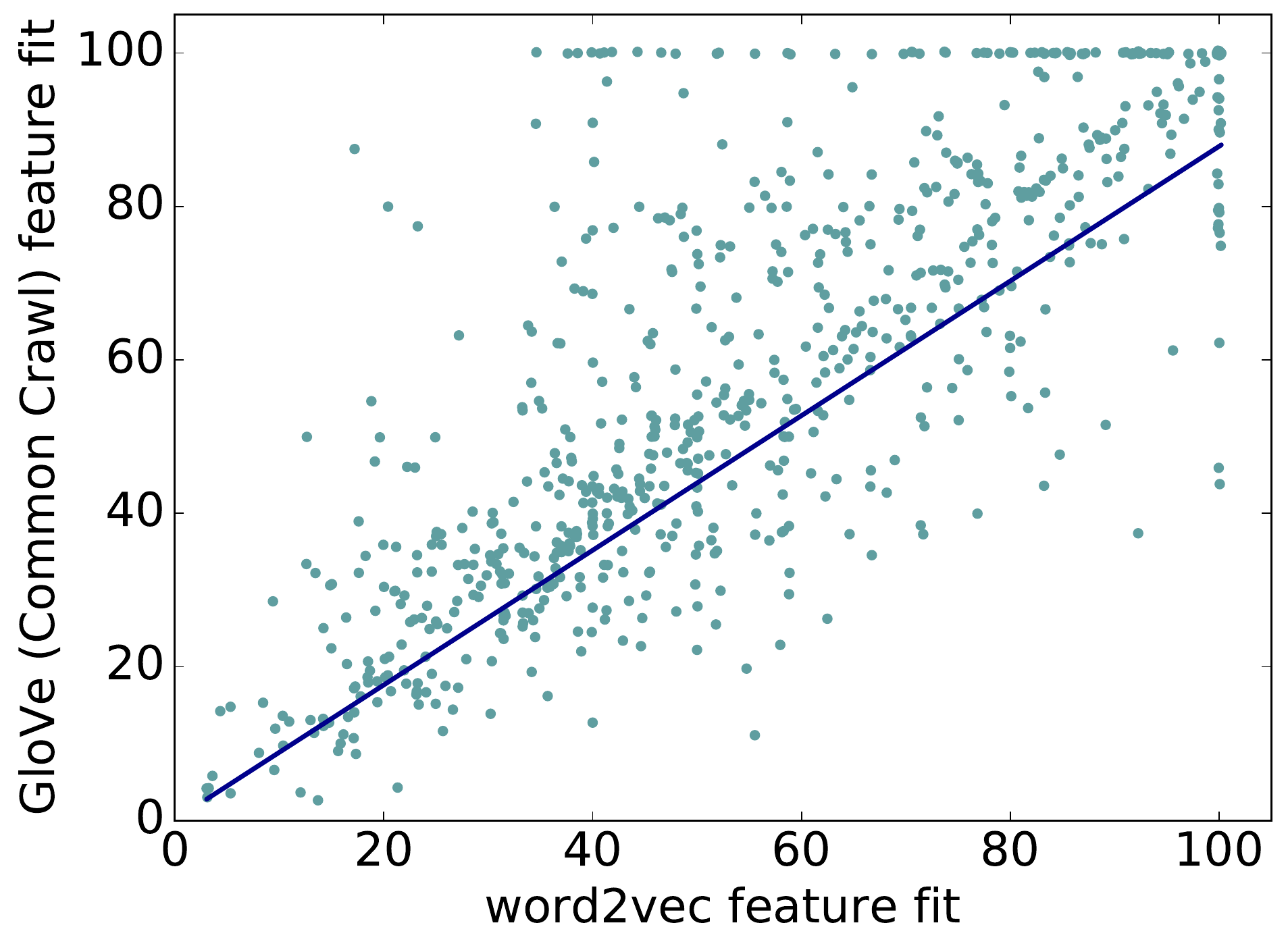}
\caption{A comparison of CSLB feature fit scores for word2vec and GloVe-CC. Slope: 0.8773; Pearson $r$: 0.8260.}
\label{fig:rep_match}
\end{figure}

\begin{figure*}[htb]
\centering
\subfloat{\includegraphics[width=0.4\linewidth]{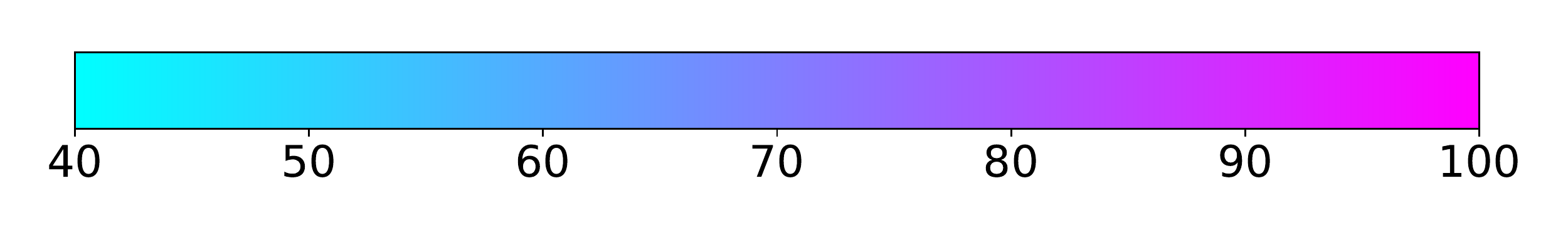}}\\[-10pt]\addtocounter{subfigure}{-1}
\subfloat[Concepts plotted according to the correlation between their GloVe-CC embeddings and two other sources (CSLB, horizontal axis; WordNet, vertical axis). The points are colored according to their feature fit.\label{fig:concept_graphs_pearson}]{\includegraphics[width=0.48\linewidth]{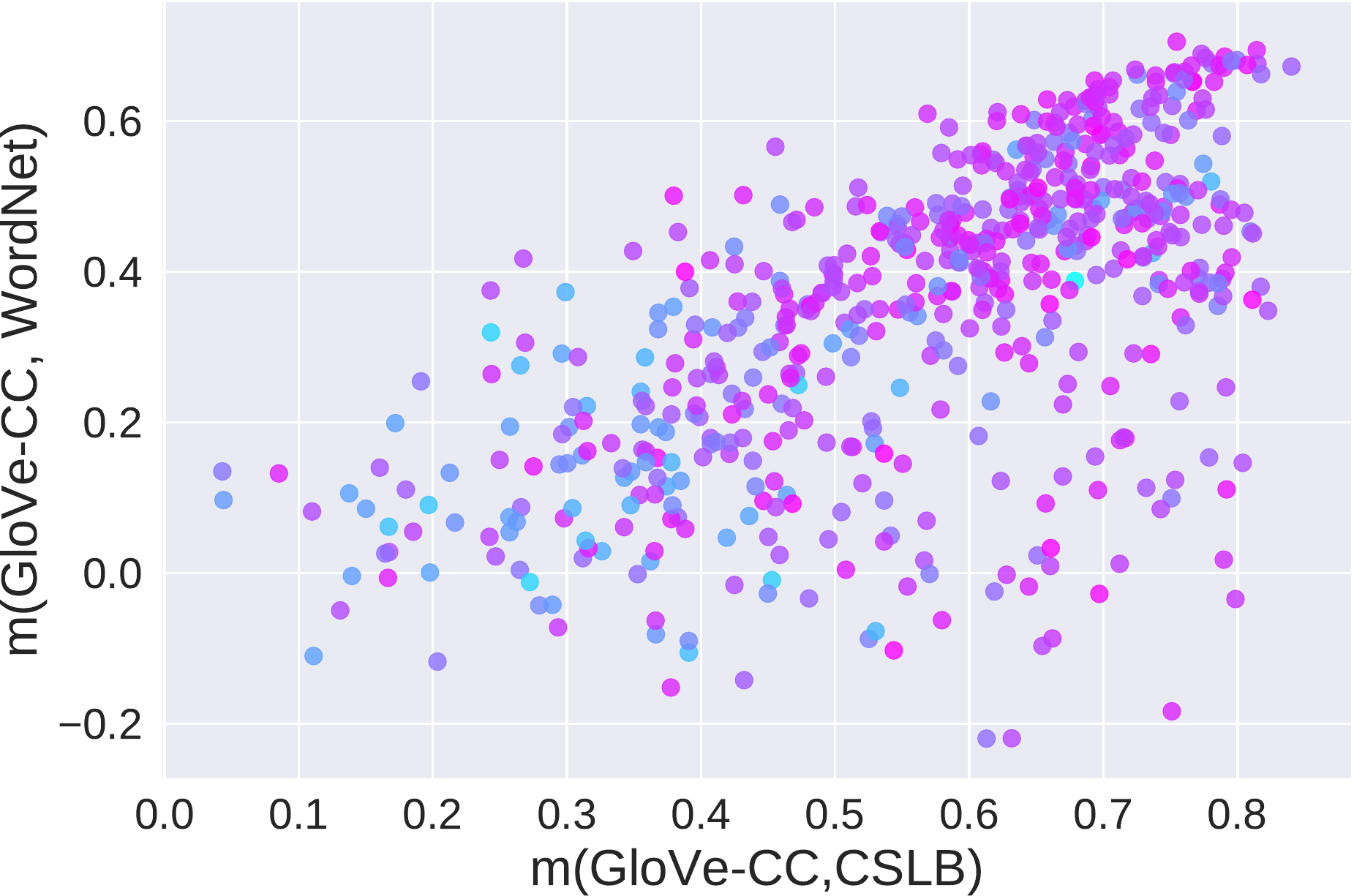}}\hfill
\subfloat[A direct comparison of $m(\text{GloVe-CC},\text{CSLB})$ (horizontal axis) and the median feature fit scores associated with concept (vertical axis). See main text for statistical tests.\label{fig:concept_graphs_ff}]{\includegraphics[width=0.48\linewidth]{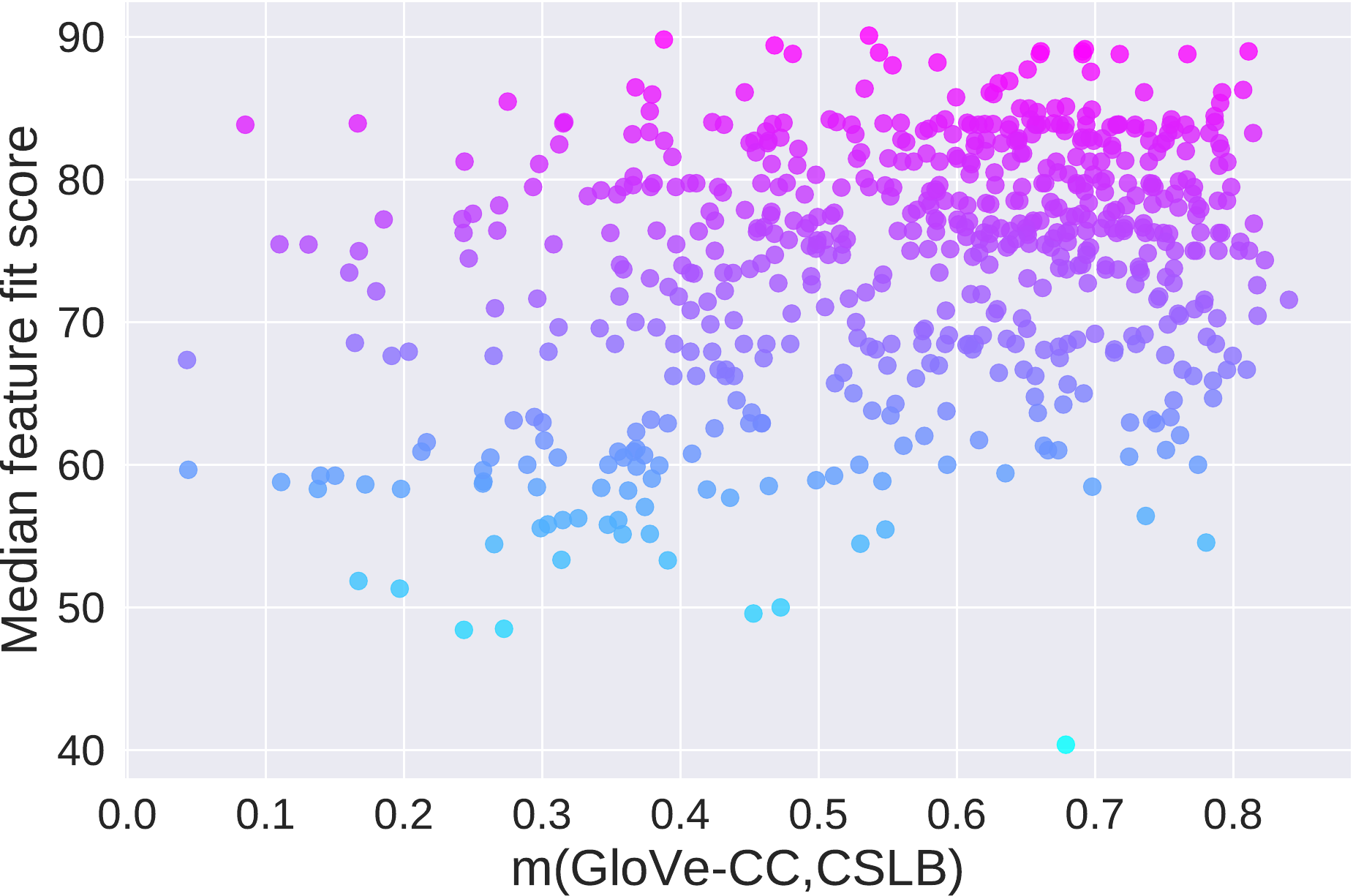}}
\caption{Concept view results.}
\label{fig:concept_graphs}
\end{figure*}

For each feature, we compare its feature fit score evaluated with GloVe-CC word vectors and its score evaluated with word2vec vectors in \Cref{fig:rep_match}. The trend in the figure suggests that both representations have similar feature fit deficiencies and strengths, though the trend becomes weaker near the $(100\%,100\%)$ corner --- the two representations correlate well at low feature fit scores, and seem to fan out at higher scores. A large group of points also sit in the figure at $y=100$ and $x=100$; these features are perfectly captured by one representation and not by the other.

This correlation is somewhat surprising, given that the word2vec and GloVe vectors are the products of different algorithms executed on very different corpora. There are two likely explanations behind this correlation:
\begin{enumerate}
\item Some features in the CSLB semantic norm data are unusually difficult, or are perhaps missing associated concepts. GloVe and word2vec correlate in performance because they don't match these noisy or incomplete features.
\item There are systematic deficiencies in the word vectors due to their shared reliance on the distributional method.
\end{enumerate}

It is difficult to differentiate these two explanations on these small semantic norm datasets, but we hope to distinguish these in the future by testing new predictions for concepts not covered in these datasets. We will return to this idea in the conclusion of the paper.

\section{The concept view} % CONCEPT VIEW

The previous section demonstrated that several classes of perceptual features are not well encoded on average by distributional word representations, and that these deficiencies systematically match across representations. How does this deficiency in feature representation carry over into computations on the word representations themselves?

We evaluate the matching between distributional representations and representations from other sources by comparing their predictions of word-word similarity. For distributional word representations, we compute word-word similarity by cosine distance:
\begin{equation}
\text{sim}(i, j) = \cos(x_i, x_j)
\end{equation}

We derive compact concept representations from the semantic norm datasets with LSA \citep{landauer1998introduction}. We compute a truncated SVD on the feature matrix $Y \in \{0,1\}^{n_c \times n_f}$, which is the concatenation of the binary feature label vectors introduced in \Cref{sec:feature-view}. We define concept-concept similarity by the cosine distance between their corresponding LSA vectors.

As a secondary data source, we also compute word-word similarity judgments from the WordNet taxonomy \citep{miller1995wordnet}. We use the Resnik metric~\citep{resnik1999semantic} to compute the similarity between concept names $c_i$, $c_j$:
\begin{equation}
\text{sim}_\text{resnik}(c_i, c_j) = \max\limits_{c \in S(c_i, c_j)} -\log p(c)
\end{equation}
where $S(c_i, c_j)$ selects the common ancestors of the concepts in the WordNet taxonomy, and $p(c)$ is the unigram probability of a concept as computed on an external corpus. This selects the ancestor of the two concepts in the taxonomy which has maximal information content (surprisal). We use WordNet as additional verification that the trends observed between semantic norms and distributional representations are non-coincidental.

\paragraph{}We use these similarity metrics to compute pairwise distance measures for concepts present in the semantic norm datasets. For each metric, we produce a symmetric pairwise distance matrix $D \in \mathbb R^{n_c \times n_c}$, where an element $D_{ij}$ indicates the distance between concepts $i$ and $j$ according to the metric.

We next compute how well each concept's pairwise similarity is correlated between the various metrics. For a given concept, we compute the Pearson correlation between the concept's GloVe/word2vec pairwise distance vector and the LSA and WordNet pairwise distance vectors.\footnote{The Pearson correlation between two vectors is equivalent to the cosine distance between their mean-centered forms.} The correlation values of interest are $m(\text{GloVe/word2vec}, \text{CSLB})$ and $m(\text{GloVe/word2vec}, \text{WordNet})$ --- that is, the correlations between the pairwise distance vectors for GloVe/word2vec and CSLB and between the pairwise distance vectors for GloVe/word2vec and WordNet.

\Cref{fig:concept_graphs_pearson} plots both of these correlation values evaluated with GloVe-CC for all concepts. The two $m$ measures are evidently positively correlated, though with some noise ($r=0.6160$). This is to be expected, as the CSLB dataset and WordNet overlap only partially in the semantic features they encode.

Each concept in \Cref{fig:concept_graphs} is colored according to the median feature fit score of its associated features. In \Cref{fig:concept_graphs_ff}, we show this feature fit metric on the vertical axis. There is a positive relationship here between feature fit scores and the correlation metric $m(\text{GloVe-CC}, \text{CSLB})$ ($r=0.3323$). Because the correlation between $m(\cdot, \text{CSLB})$ and feature fit metrics is weaker than expected, we run post-hoc multiple regression significance tests for each distributional representation. An F-test shows that the regression feature $m(\cdot, \text{CSLB})$ significantly improves predictions of feature fit values relative to a baseline model for all three representations\footnote{The baseline regression model predicts a concept's feature fit from these baseline features: $\log$(word frequency in Brown corpus), $\log$(\# associated features), $\log$(total \# feature reports for the concept), \# WordNet senses.}\footnote{GloVe-CC: $F^*=41.297,p<10^{-9}$; GloVe-WG: $F^*=68.783,p<10^{-15}$, word2vec: $F^*=41.27,p<10^{-9}$}.

There is substantial variance in the predictions of the distributional representations due to factors outside of the scope of the semantic norm data. The mismatch in predictions between distributional representations is nevertheless a statistically significant predictor of feature fit metrics. This suggests that the feature-level deficiencies discovered in the previous section have concrete implications in terms of word-word similarity measures.

\subsection{Domain-level analysis} % DOMAINS

\begin{figure*}[t]
\centering
\includegraphics[width=0.85\linewidth]{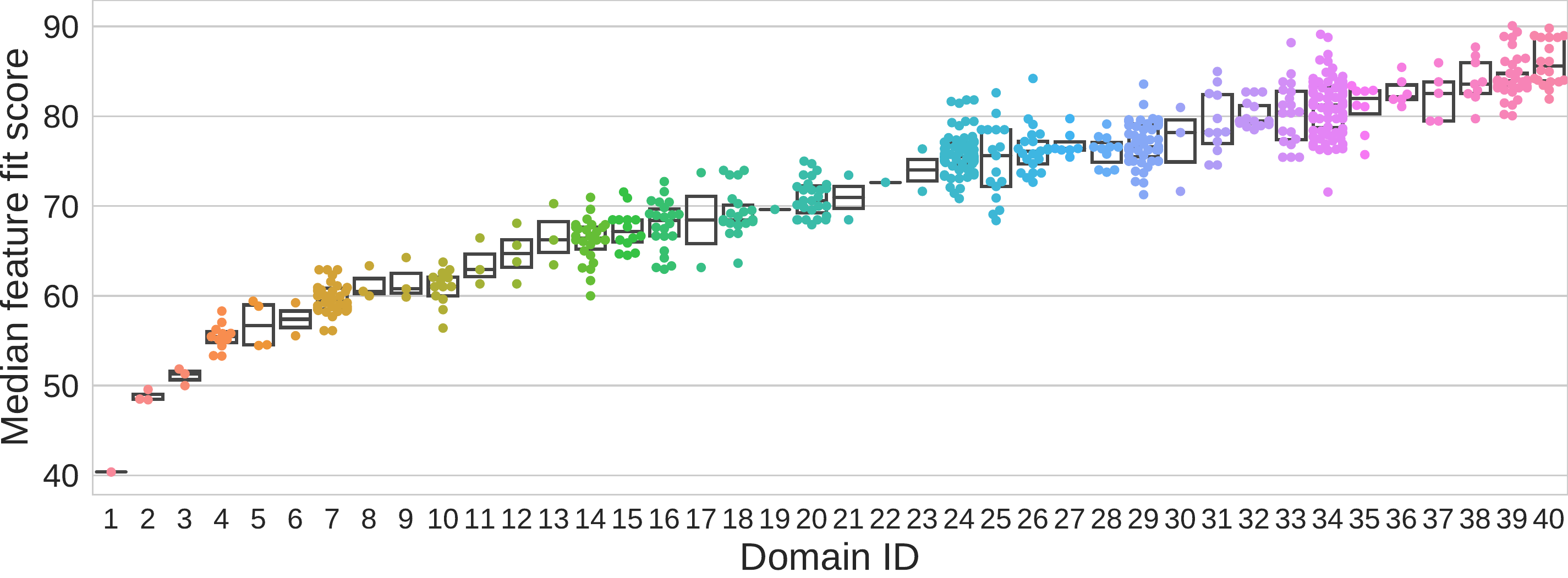}
\caption{Concept domains derived from the CSLB semantic norm data. Each point represents a concept. The vertical axis is the median feature fit score of the concept's features on GloVe-CC.}
\label{fig:domain_box}
\end{figure*}

\begin{table}[t]
\centering
\resizebox{\linewidth}{!}{
\begin{tabular}{l|l|p{6cm}}
\toprule
Domain & Feature fit & Concepts \\
\midrule
10 & 61.03\% & \nohyphens{bread, cheese, chocolate, coffee, glue, ham, jam, jelly, ketchup, moss, soup, tea, yoghurt}\\
15 & 67.18\% & \nohyphens{artichoke, asparagus, aubergine, bean, cabbage, flour, gherkin, leek, mango, pineapple, potato, pumpkin, rhubarb}\\%, seaweed}\\
25 & 75.62\% & \nohyphens{bouquet, buttercup, carnation, daffodil, daisy, dandelion, fern, geranium, hyacinth, lily, marigold, orchid, pansy, poppy, rose, sunflower, tulip}\\
31 & 78.22\% & \nohyphens{bayonet, bomb, cannon, crossbow, dagger, grenade, gun, pistol, revolver, rifle, shotgun, sword}\\
36 & 82.18\% & \nohyphens{book, catalogue, menu, dictionary, encyclopaedia, textbook}\\
\bottomrule
\end{tabular}
}
\caption{Selected domains from the clustering analysis on GloVe-CC, with median feature fit scores over concepts.}
\label{tbl:domains_feature_fit}
\end{table}

We next investigate whether some domains of concepts are particularly affected by the deficiencies discussed in the previous sections. We perform agglomerative clustering on concepts from the CSLB dataset using a custom distance metric:
\begin{equation}
d(i, j)=||\text{LSA}_i - \text{LSA}_j||_2 + \alpha (\text{FF}_i - \text{FF}_j)^2
\end{equation}
where $\text{LSA}_i$ is the LSA vector representation computed from the semantic norm data for concept $i$ as introduced earlier in this section, and $\text{FF}_i$ is the median feature-fit score for a concept $i$. We select the weight $\alpha$ manually to produce the most semantically coherent clusters.

\Cref{fig:domain_box} shows the distribution of feature fit scores for each of the resulting 40 domains. We find that settings of $\alpha$ which yield semantically coherent clusters also yield groups of concepts with very low variance in feature fit scores. In \Cref{tbl:domains_feature_fit} we list select domains and their median feature fit scores. This clustering suggests that deficiencies at the feature level affect entire coherent semantic domains of concepts.

\section{Conclusion}

This paper has analyzed how well various standard distributional representations encode aspects of grounded meaning. We chose to use semantic norm datasets as a gold standard of grounded meaning, and tested how word representations predicted features within these datasets. We grouped these features into high-level categories and found that, despite large within-category variance, several standard distributional representations underperformed on average in predicting perceptual features. The difference in prediction performance proved statistically significant on two of the three representations we evaluated. These deficiencies in feature encoding matched between GloVe and word2vec representations trained on different corpora, suggesting that certain classes of features may be poorly represented by distributional methods in general.

We also examined the consequences of these deficiencies in feature encoding for the word representations themselves. We compared the word-word similarity predictions made with distributional representations with those made with the semantic norm dataset and with WordNet, and found that words having features badly encoded within the distributional representations were also likely to make different similarity predictions than the predictions from these two corpora. A final domain-level concept analysis suggested that some semantic domains are particularly impacted by these issues in feature encoding.

\paragraph{}The semantic norm datasets used in this paper are subject to saliency biases: they only contain the concept-feature mappings which experimental subjects think to mention when queried. These saliency effects add noise to our results, as mentioned in \Cref{sec:feature-comp}, and may have caused us to generally underestimate the performance of distributional models within all feature categories. In future work, we plan to repeat the sorts of tests conducted in this paper while avoiding possible saliency confounds. We also plan to develop a causal explanation for the deficiencies in the word embeddings found in this paper, showing how co-occurrence information (or lack thereof) present in the training corpus can bias performance on these tasks. Both of these studies will verify that the results we have found are due entirely to deficiencies in distributional methods rather than in the datasets used here.

\paragraph{}We think these deficiencies should be worrying: if neural models of language are to have any knowledge about concepts, it ought to be in their word embeddings. Our findings show that these embeddings are lacking in basic features of perceptual meaning. These results suggest that distributional meaning (as operationalized by modern distributional models) may miss out on fundamental elements of semantics. We hope they will help motivate further work in developing multimodal representations which can prepare us to deploy more fluent language agents in the real world.

\section*{Acknowledgements}
We thank Christopher D. Manning, Peng Qi, Pakapol Supaniratisai, Keenon Werling, and members of the Stanford, University of Washington, and Berkeley NLP communities for useful discussions, and the anonymous reviewers for their insightful comments.

\bibliography{acl2017}
\bibliographystyle{acl_natbib}

\end{document}